# Fake News Detection and Behavioral Analysis: Case of COVID-19

**Chih-Yuan Li[1], Navya Martin Kollapally[2], Soon Ae Chun[3], James Geller[4]**
[1,2,4]New Jersey Institute of Technology, [3]College of Staten Island, City University of New York
{[1]cl524, [2]nk495, [4]james.geller}@njit.edu, [3]soonaechun@gmail.com

## Abstract

While the world has been combating COVID-19 for over three years, an ongoing "Infodemic" due to the spread of fake news regarding the pandemic has also been a global issue. The existence of the fake news impact different aspect of our daily lives, including politics, public health, economic activities, etc. Readers could mistake fake news for real news, and consequently have less access to authentic information. This phenomenon will likely cause confusion of citizens and conflicts in society. Currently, there are major challenges in fake news research. It is challenging to accurately identify fake news data in social media posts. In-time human identification is infeasible as the amount of the fake news data is overwhelming. Besides, topics discussed in fake news are hard to identify due to their similarity to real news.

The goal of this paper is to identify fake news on social media to help stop the spread. We present Deep Learning approaches and an ensemble approach for fake news detection. Our detection models achieved higher accuracy than previous studies. The ensemble approach further improved the detection performance. We discovered feature differences between fake news and real news items. When we added them into the sentence embeddings, we found that they affected the model performance. We applied a hybrid method and built models for recognizing topics from posts. We found half of the identified topics were overlapping in fake news and real news, which could increase confusion in the population.

## 1. Introduction

### 1.1 Motivation

Social media users share and are exposed to substantial amounts of information through social media platforms. Although it has become easier to retrieve information, it has also become harder to verify the authenticity of the information. Because of the convenience of accessing social media and sharing posts, many users might distribute fake news due to evil intentions, ignorance, or for personal gain or entertainment. While the COVID-19 pandemic has led to an unprecedented loss of lives and economic disruptions, fake news on COVID-19 has also become an issue. For example, some people have accepted mistaken reports that garlic, alcohol, or vitamins could prevent and/or cure COVID-19. This in turn might have led them to ignore warnings about wearing masks and staying socially distanced, potentially leading to their infection, hospitalization, or even death. Fake news has also claimed that COVID-19 is a hoax, not worse than the “normal” flu, and vaccinations are not effective or even dangerous. This flood of fake news has been termed an “Infodemic” [1], i.e., the appearance of too much misleading information during a disease outbreak.

In this paper, our goal is to use Machine Learning to find remedies for this Infodemic. The objectives of this paper are the following. We aimed to provide Deep Learning-based detection models for differentiating fake news from real news. We built topic identification models to retrieve topics appearing in fake news and in real news regarding COVID-19. We identified feature differences between fake news and real news items about the pandemic.

In order to help curb this "health information disorder," [2] we need to identify fake news first. We present a Deep Learning approach to distinguish between fake news and real news. We used LSTM [3] , BERT [4], and DistilBERT [5] as our Deep Learning algorithms, and SVM [6] for creating an ensemble model. To train Deep Learning models, we utilized three publicly available fake news datasets. Two of them were news-based, and were collected before the COVID-19 pandemic. The third one was a set of social media posts of fake news and real news on COVID-19, which is our main focus. The SVM model was built specifically for the third dataset.

To further realize the features of fake news posts generated and shared on social media amid the COVID-19 pandemic, we applied Natural Language Processing (NLP) techniques, and implemented behavioral and sentiment analyses focusing on the third dataset about COVID-19. We use the term "behavioral features" for features that are directly the result of a choice by the user, e.g., whether to use a "#hashtag" for a certain word, or to provide a hint about how the user feels about an issue ("concern indices"). Several behavioral features were analyzed and compared between fake news and real news, including length, sentiments, "concern indices," and the use of hashtags (e.g., #COVID) and mentions (e.g., @WHO). Then, to investigate how influential the features could be toward fake news detection, we built Deep Learning models based on feature elimination. We also built an ensemble model with the identified features, combining BERT and SVM, to achieve improved performance.

Moreover, we built topic identification models to identify the topics manipulated in fake news and the topics discussed in real news about COVID-19. A number of interesting topics were identified, half of which were common between fake news and real news. Our experiment also showed that fake news has a relatively more consistent writing style.

To better distinguish between fake news and real news regarding COVID-19, we are raising five research questions in this paper: (1) Is there a difference of the expressed sentiments between fake news and real news of social media posts regarding COVID-19? Besides, if such a difference exists, is it statistically significant? (2) What are the differences between fake news and real news with regards to easily measurable features, such as number of words and number and use of hashtags and mentions in a posting, and can these features contribute to better accuracy in fake news detection? (3) Are there differences in the topic structure of fake news compared to real news, and what are the main topics for COVID-19 in both? (4) How good are the results when we use transfer learning with models trained with fake news datasets that predate COVID-19 when tested with a COVID-19 fake news dataset? (5) Do the extracted features help improve model robustness?

### 1.2 Contributions

We have achieved several contributions in this paper, which include:

1. Our fake news detection models achieved better performances than previous studies.
2. We found that the identified behavioral features can be used to distinguish between fake news and real news.
3. We found that half of the identified topics are common in fake news and real news. This shows that the existence of fake news could make readers confused and make them mistake fake news for real news.
4. We found that fake news causes significantly more concerns among citizens, and the "concern

indices" of fake news and real news are significantly different.
5. We performed ensemble modeling, and we further improved the detection model's accuracy.

### 1.3 Paper Overview

This paper is organized as follows. Section 2 describes related work. In Section 3 we present our datasets and the Machine Learning methods used for our research, as well as the topic modeling and sentiment analysis techniques used. Section 4 presents our results. The discussion of our experimental results appears in Section 5. We conclude this paper in Section 6, and briefly describe our ongoing research in Section 7. A preliminary version of this work appeared as a short paper in the FLAIRS conference [7].

## 2. Related Work

Even though "fake news" has become common, there is still no agreement on its definition. Allcott et al. (2017) [8] defined fake news as "news articles that are intentionally and verifiably false and could mislead readers." Lazer et al. (2018) [9] defined fake news as "fabricated information that mimics news media content." Another definition [10] says that fake news "is false or misleading information presented as news. It often has the aim of damaging the reputation of a person or entity, or making money through advertising revenue." Some prior work saw satire news as fake news since the contents are false, even though satire is often entertaining and reveals its own purpose to the readers [11][12][13][14]. Although the definitions vary, it is important to recognize that they all exclude unintentional reporting mistakes [15][16][17][18].

Trusting fake news or misleading contents can cost lives. In March 2020 in Iran, nearly 300 people died and more than 1000 were sickened after ingesting methanol, because of a fake news message that "alcohol can wash and sanitize the digestive system" [19]. In the study of [20], the author presented COVID-19 fake news items, including "transmission via mosquito bites," "temperature as a cure," "youthful immunity," etc. Such information might cause fatalities directly, and might make the situation worse by not seeking out legitimate remedies. Fake news can also cause street riots. In Feb 2020 in the central Ukrainian town of Novi Sanzhary, a fake news email, purportedly from the Ministry of Health, provoked panic by claiming that some of the evacuees from the town were infected with the virus. This led to local residents burning tires, blocking roads, and clashing with the hundreds of policemen who were urgently dispatched to disperse them [21].

In order to reduce the impact of fake news, we need to identify such false information first. Supervised Machine Learning algorithms such as Decision Tree, Random Forest, Support Vector Machine, Logistic Regression, K-nearest Neighbors are extensively used for fake news detection [22][23][24][25][26][27][28]. Bojjireddy et al. (2021) [29] presented a Machine Learning approach to recognizing misinformation, specifically using presidential election and COVID-19 related fake news [35][41][62][63][65]. They applied several Machine Learning approaches – Multinomial Naïve Bayes (MNB), Support Vector Machine (SVM), Multilayer Perceptron, Decision Tree, Random Forest, and Gradient Boosting (GB) as technical solutions for automating the detection of fake news and misleading contents.

In recent years, Deep Learning has gained a favorable reputation in the areas of speech recognition and visual object recognition [30][31][32]. Machine Learning techniques require humans to reduce the complexity of the data and make patterns more visible (e.g., by adding combined features) for algorithms to work well. Deep Learning algorithms can be fed with raw data and they discover the representations, without the need for human feature engineering [31][33]. Ali et al. (2021) [34] investigated the robustness of different Deep Learning architecture choices, e.g., Multilayer Perceptron (MLP), Convolutional Neural Networks (CNN), Recurrent Neural Networks (RNN) and a recently proposed Hybrid CNN-RNN combi-nation. Their experiments based on the Kaggle fake-news dataset [35], ISOT dataset [36], and LIAR dataset [37] suggest that RNNs are robust as compared to other architectures. Li et al. (2021) [38] proposed an unsupervised method based on an autoencoder to detect fake news in the MediaEval 2016 fake news dataset [39]. The detection was based on features including text content, images, propagation information, and user information (followers, likes) of published news.

In this paper, we built several Deep Learning approaches for fake news detection. Using Deep Learning models eliminates the need for domain expertise and hard-core feature extraction [33]. Second, Deep Learning outperforms other techniques if the data size is large. Third, when there is a lack of domain understanding for feature introspection, Deep Learning stands out, as there is less worry about the need for feature engineering with it [40].

The availability of curated fake news datasets is one of the fundamental factors for building an effective fake news detection model. Currently there exists a number of available fake news datasets: Benjamin Political Dataset [41][42], Burfoot Satire News Dataset [43], BuzzFeed News [44], Credbank Dataset [45], Fake News Challenge Dataset [46], FakeNewsNet [47], etc.

In order to deal with the polarization of society over controversial social issues, Matakos et al. (2017) [48] used a standard opinion formation model to capture the tendency of opinions to concentrate in network communities, creating echo-chambers. As the existence of fake news will likely cause further confusion of citizens and conflicts in society [49][50], De Choudhury et al. (2013) [51] developed a probabilistic model that leverages signals of social activity, emotions, and language manifested on Twitter to determine if posts could indicate depression. In this paper, we use a "concern index" metric to measure the concerns expressed by sets of posts. We further compute the significance of concern index differences between fake news and real news. Since the outbreak of the COVID-19 pandemic, the U.S. Federal government has started considering introducing legislation to make it an offence to spread harmful misinformation [52].

## 3. Methods

In this section, we present our methods used in this paper. Figure 1 shows the pipeline of the work. We will discuss the methods and processes in detail in the following subsections.

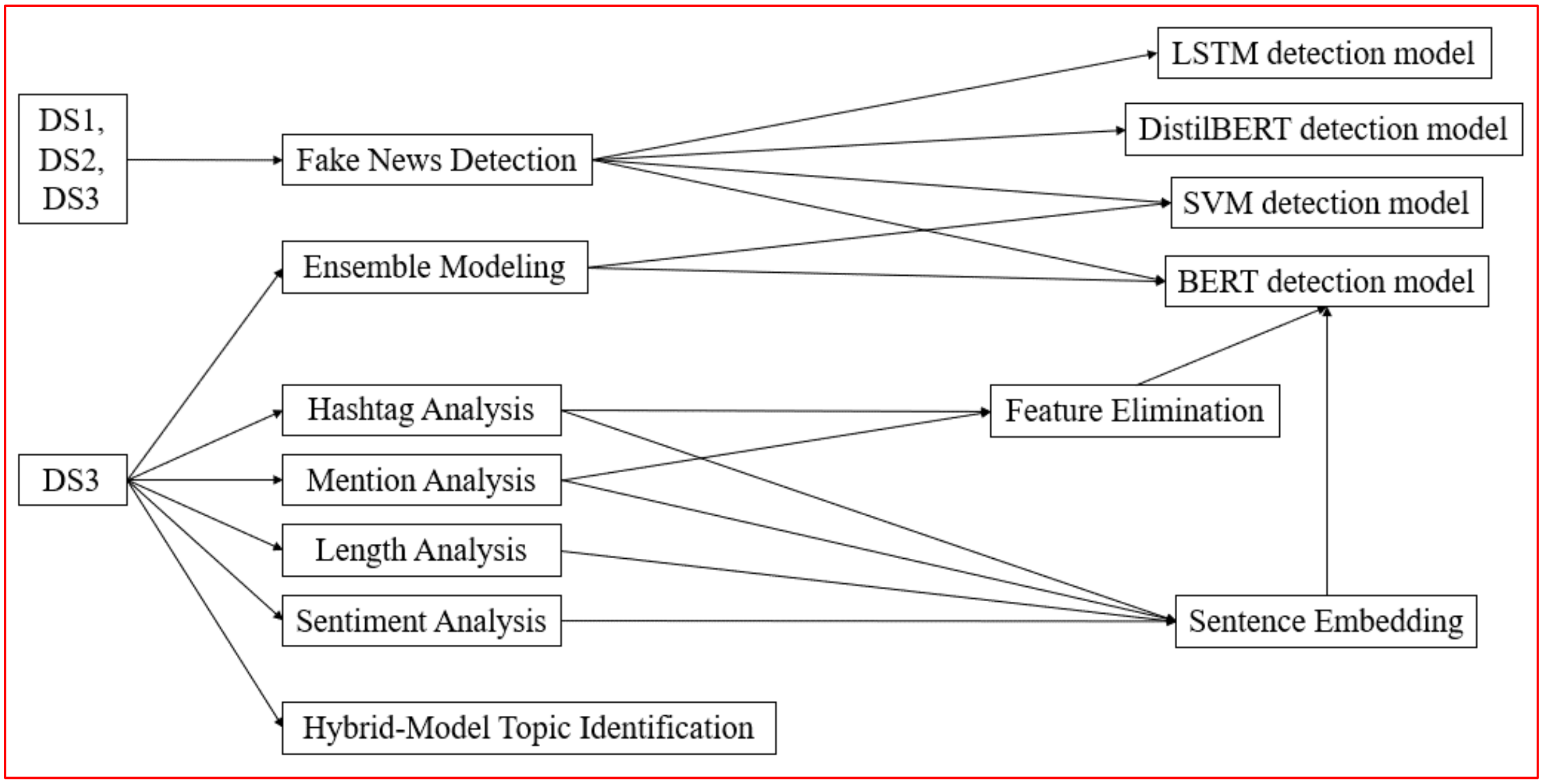

Figure 1: Data processing pipeline.

### 3.1. Fake News Detection Models

For fake news detection, we built SVM [6], LSTM [3], BERT [4], DistilBERT [5], and models.

SVM (Support Vector Machine) is a supervised learning model for two-group classification problems. SVM performs linear classification, and non-linear classification by using the kernel trick [53]. We used an RBF kernel. SVM works effectively in high-dimensional spaces. It is also effective when the number of dimensions is greater than the number of samples.

LSTM (long short term memory) is a specific recurrent neural network (RNN) that can handle long term dependencies and in turn solve the problem of vanishing gradient. A common LSTM unit is composed of a cell, an input gate, an output gate and a forget gate. The cell remembers values over arbitrary time intervals, and the three gates regulate the flow of information into and out of the cell.

BERT is a powerful Deep Learning system for language modelling, and is the first deeply bidirectional model. It uses bidirectional transformers, such that a transformer is used for converting a sequence using an encoder and a decoder into another sequence. As opposed to directional models, which read the input text sequentially (left to right or right to left), the transformer encoder reads the entire input text at once. This characteristic allows the model to learn the context of a word based on its left and right neighboring words.

DistilBERT is a pre-trained version of BERT. DistilBERT leverages knowledge distillation during a pretraining phase. Thus, DistilBERT has fewer parameters than the Bert model (bert-base-uncased) by 40%, while it retains 97% of its language understanding capabilities and runs 60% faster [5]. From BERT the token-type embeddings and the poolers are removed while the number of layers is reduced by a factor of 2.

For each model, we performed 5-fold cross validation. In this paper, we used three datasets, and trained each of the LSTM, BERT, DistilBERT models with each dataset. Therefore, we will describe nine training regimens.

### 3.2. Topic Identification

In this Section, we present our methods used to discover topics discussed and mentioned in DS3. The purpose why we performed topic identification is as follows. We would like to explore the topics discussed in real news and in fake news. If there are many common topics in both, they would be more likely to confuse readers, because the readers would mistake fake news for real news. We compared the topics in fake news posts with those in real news posts. During text preprocessing for this task, for each input news item, we removed stop words, numbers, and punctuations, lowercased each letter, and lemmatized each word. We then applied a hybrid method by using Latent Dirichlet Allocation (LDA) [54] and BERT [4]. LDA represents documents as random mixtures over latent topics, where each topic is characterized by a distribution over words [54]. It is effective for identifying topics by finding frequent words when sentences are coherent. However, when text is not coherent, extra contextual information is needed to comprehensively represent the idea of a text item. As social media posts often do not form complete sentences, LDA needs to be augmented by another mechanism. The transformer encoder of BERT reads the entire sequence of words at a time. This characteristic allows the model to learn the context of a word based on its surroundings (left and right of the word). By combining the probabilistic topic assignment vector from LDA with the sentence embedding vector from BERT, we can augment semantic information with contextual topic information.

The concatenated vectors of LDA and BERT will yield high-dimensional spaces, which arise as a way of modeling datasets with many features. However, the number of features can exceed the number of observations, and the calculations become difficult. To deal with this issue, an autoencoder [55] is used to learn a lower dimensional latent space representation of the concatenated vector. The latent space is a representation of compressed data in which similar data points are closer together in space. The low dimensional latent space aims to capture the most important features required to learn and represent the input data. We implemented K-Means clustering [56] on the latent space representations, and we assigned contextual topics to the clusters. To decide on the best number of topic clusters, we applied the *coherence score* metric [57]. Coherence score is a measure of scoring a single topic by measuring the degree of semantic similarity between words in each topic. A higher coherence score indicates a higher degree of likeness in the meaning of the words within each topic.

### 3.3. Extraction of Hashtags and Mentions

We used Python regular expressions to capture any word starting with a “#” (hashtag), or a “@” (mention of a Twitter user). There were hashtags expressing the same meaning, but in different representations, such as “Covid_19” and “covid19.” We lowercased each hashtag and removed the punctuations (Table 1). There are two purposes for the extraction of hashtags and mentions. First, we would like to see whether they are used differently in fake news and real news. Second, we aim to improve the accuracy of detection models by adding hashtag counts and mention counts as new features.

Table 1: Examples of hashtag cleaning.

| Original Hashtag | Processed Hashtag |
|---|---|
| Covid_19 | covid19 |
| CoronaVirusFacts | coronavirusfacts |
| NYCLockdown | nyclockdown |

### 3.4. Sentiment Analysis

We measured the sentiments of the news items to determine their emotional impact. There are a number of sentiment analyzers in the literature. The analyzer we utilized in this paper is from the Stanford NLP library [58]. The Stanford Sentiment Analyzer uses a fine-grained analysis based on both words and labeled phrasal parse trees to train a Recursive Neural Tensor Network (RNTN) model [59]. RNTN represents a phrase through word vectors and a binary parse tree. RNTN computes parent vectors in a bottom-up fashion using a compositionality function, using node vectors as features for a classifier at each node. In other words, the RNTN model computes the sentiment expressed by a sentence, based on how words compose the meaning of longer phrases. An input phrase is labeled either as “Very Negative,” “Negative,” “Neutral,” “Positive,” or “Very Positive.” The source code is available at [60]. Figure 2 shows an example of the compositional structure predicting the sentiment class for each node in the parse tree. Table 2 shows post samples with labelled sentiments.

To monitor the concerns expressed by fake news and real news, we then computed a “Concern Index,” following Ji et al. (2013) [61]. The higher the Concern Index is, the bigger the negative sentiment that is expressed by the news item. There are three reasons for performing sentiment analysis. First, we derive the sentiment distribution from the posts. Second, we use the distribution to calculate the concern indices and identify whether the difference of concern indices between fake news and real news is significant. Third we add the numerical values of the sentiments as new features to the training data to improve the detection model accuracy.

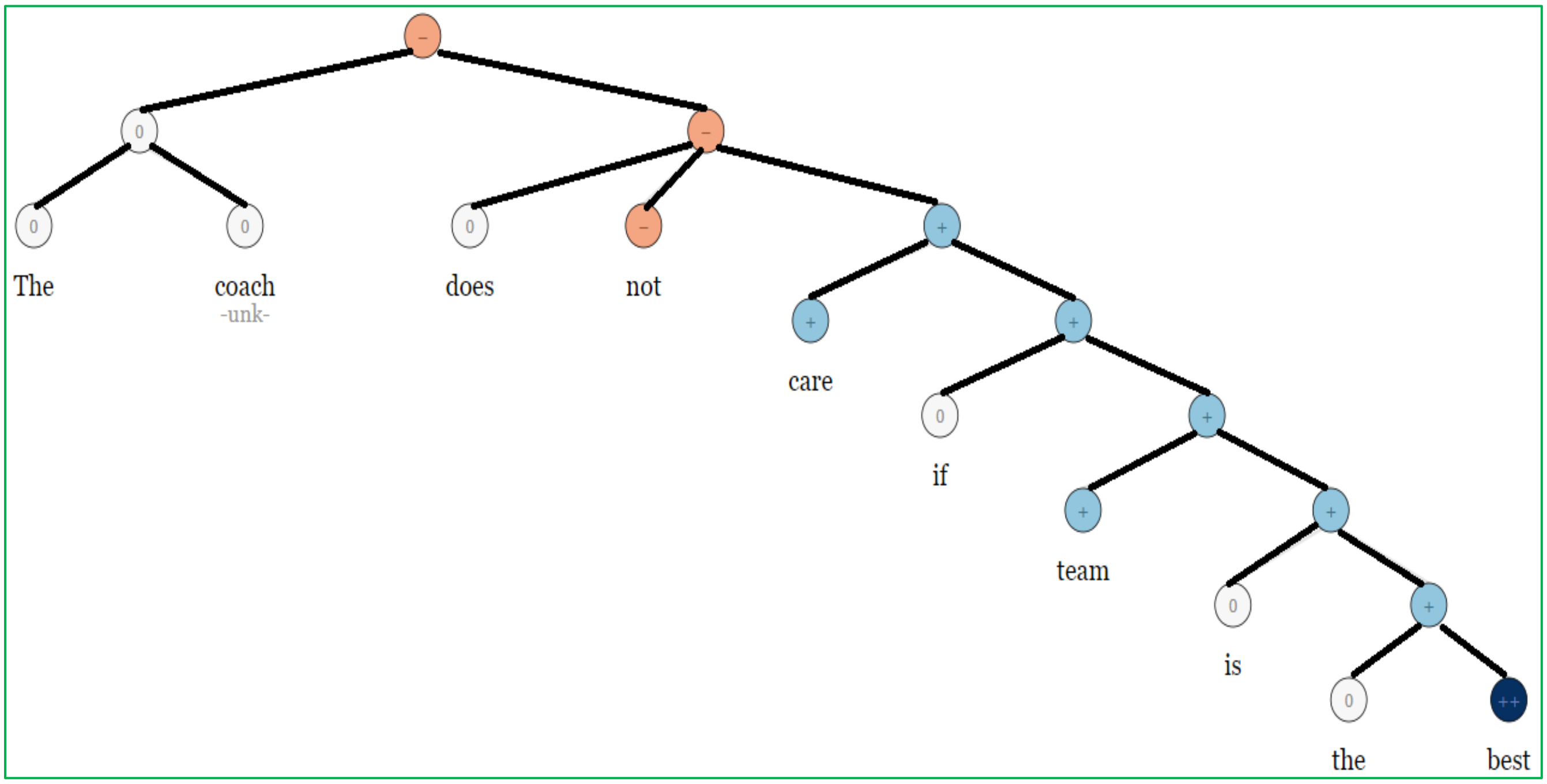

Figure 2: An example of the RNTN predicting the sentiment classes, Negative (-), Neutral (0), Positive (+) and Very Positive (++), as well as capturing the negation in the sentence [59].

**Definition 1.** Concern Index calculation.

$$\text{Concern Index} = \frac{N}{N+P+1} \tag{1}$$

N is the count of items with *Negative* sentiments + the items with *Very Negative* sentiments. P is the count of items with *Positive* sentiments + the items with *Very Positive* sentiments. We are purposefully not using the items classified as *Neutral* for computing the Concern Index. As opposed to Ji et al. (2013), we add 1 to the denominator to avoid the (unlikely) case of division by zero.

Table 2: Examples of the five sentiment classes, taken from the dataset DS3.

| Text | Sentiment |
|---|---|
| A voice recording on WhatsApp claims there will be 900 deaths a day at the peak of the new coronavirus outbreak, one third will be children with no underlying health conditions, ambulances won't be sent to people struggling to breathe, and all ice rinks are now being used as mor... | Very Negative |
| Obama Calls Trumps Coronavirus Response A Chaotic Disaster | Negative |
| In May we did not break 30k cases in a day. Today the South alone reported 32830. | Negative |
| India spraying pesticides at night to prevent COVID19. | Neutral |
| Our daily update is published. States reported 586k tests 28k cases and 224 deaths. | Neutral |
| It is best to shave your beard to avoid being infected by the coronavirus. | Positive |
| For example once a successful vaccine has been identified WHOs strategic advisory group will provide recommendations for their appropriate and fair use. The allocation of vaccines is proposed to be rolled out in two phases- @DrTedros #COVID19 | Positive |
| foundation is truly one of the most inspirational forces of social change | Very Positive |

## 4. Experiments and Results

### 4.1 Dataset

To create a large training dataset, we combined multiple sources of publicly available fake news data.

*Dataset 1 & Dataset 2.* Data items collected from a political dataset [41], a Kaggle dataset [35], fakeorreal [62], and from Kaggle fakereal [63] contain both news titles and news content text of fake news and real news. Data items collected from Politifact [64] and [65] contain only news titles of fake news and real news. These data items were collected before the COVID-19 pandemic. We grouped the news titles together as dataset 1 (DS1), and the news content text items as dataset 2 (DS2). Therefore, DS1 includes news titles, with 69,027 fake news items and 84,232 real news items. DS2 includes news content text items, with 37,115 fake news and 66,067 real news items.

*Dataset 3.* As our main interest in this paper is in fake news about COVID-19, we used a third dataset, published by [66], that contains 4,480 real news and 4,080 fake news items. Fake news items were collected from Facebook and Instagram posts, tweets, public statements, and press releases. They were verified as fake news by various fact-checking sites ([64], [67], [68]), and by tools such as Google fact-check-explorer [69], and International Fact-Checking Network [70]. These sites present determinations on posts about COVID-19 and other generic topics, whether the items are fake or real. The real news items were from Twitter, using official and verified Twitter handles, including WHO (World Health Organization), CDC (Centers for Disease Control and Prevention), ICMR (Indian Council of Medical Research), etc. Each post was read by a human, and was marked as real news if it contained useful information on COVID-19. We call this dataset "DS3." Fake news and real news examples are shown in Table 3.

Table 3: Examples of fake news and real news from DS3.

| Text | Label |
|---|---|
| Florida Governor Ron DeSantis Botches COVID-19 Response - By banning Corona beer in order to flatten pandemic curve. | Fake |
| Masks can help prevent the spread of when they are widely used in public. When you wear a mask you can help protect those around you. When others wear one they can help protect people around them incl. you. | Real |
| Americans With Coronavirus Symptoms Are Being Asked To Cough Directly Onto President Trump | Fake |
| The main mode of transmission of is through droplets and it is possible that infected smokers may blow droplets carrying the virus when they exhale. Regardless of you should steer clear of second-hand smoke as it may cause various health problems. | Real |
| Justin Trudeau Resigns Amidst Coronavirus Pandemic | Fake |
| The positive rate has fallen a lot since early April. Back then it was ~20%. Now it's more like 4-5-6%. A lot of that change has been driven by the rising tests and plummeting positive rates in the northeast. | Real |
| Experts Call Out Claims That Cow Dung/Urine, Yoga, AYUSH Can Prevent Or Treat COVID-19 | Fake |
| Georgia is particularly worrisome. The state had not seen a large rise in reported deaths despite rising infections and a steep hospitalization curve. Today the state reported its second-highest deaths since the beginning of the pandemic and the highest number since April 7. | Real |

### 4.2 Detection Models

We used three datasets and trained each of the LSTM, BERT, DistilBERT models with each dataset. Therefore, there are nine training regimens. The performance results of our Deep Learning models are shown in Table 4. In our experiments, fake news detection with BERT, which makes use of context, outperformed the other models on all three datasets. Compared with previous approaches, our BERT models, BERT1 based on DS1 and BERT3 based on DS3 achieved higher accuracy than the models by [29] and [66], respectively, as presented below in Table 5 and Table 6.

Table 4: Performance of our trained Deep Learning models.

| Model Accuracy | LSTM | BERT | DistilBERT |
|---|---:|---:|---:|
| DS1 | 86.64% | **93.5%** | 70.4% |
| DS2 | 89.29% | **97.05%** | 81.5% |
| DS3 | 87.21% | **95.61%** | 75.37% |

Table 5: Accuracy of BERT1 and previous approaches.

| **Model** (BO = Bojjireddy et al., 2021) | **Accuracy** |
|---|---|
| Multinomial Naïve Bayes (BO) | 78.87% |
| Gradient Boosting (BO) | 79.72% |
| Decision Tree (BO) | 81.58% |
| Random Forest (BO) | 86.39% |
| Our LSTM | 86.64% |
| Support Vector Machine (BO) | 87.35% |
| Multilayer Perceptron (BO) | 87.75% |
| **Our BERT1** | **93.5%** |

Table 6: Accuracy of BERT3 and previous approaches.

| **Model** | **Accuracy** |
|---|---|
| Decision Tree (Patwa et al., 2021) | 85.23% |
| Gradient Boost (Patwa et al., 2021) | 86.82% |
| Our LSTM | 87.21% |
| Logistic Regression (Patwa et al., 2021) | 92.76% |
| Support Vector Machine (Patwa et al., 2021) | 93.46% |
| **Our BERT3** | **95.61%** |

Table 7: Performance of our trained BERT models.

| Model Accuracy | DS1 | DS2 | DS3 |
|---|---|---|---|
| BERT1 = BERT trained on DS1 | **93.52%** | 73.74% | 42.65% |
| BERT2 = BERT trained on DS2 | 39.15% | **97.05%** | 46.5% |
| BERT3 = BERT trained on DS3 | 48.27% | 38.73% | **95.61%** |

To investigate the possibility of transfer learning, we evaluated our BERT models on the other two datasets that they were not trained on (Table 7, the non-diagonal values). Each of the three models was trained with DS1, DS2, and DS3. We used pre-trained BERT, and each model was fully trained with embeddings and classification. However, the experiments show that the models were not adapting well between different domains, because the datasets are intrinsically different in their nature. DS1 and DS2 are news-based and from the time before COVID-19, while DS3 is social network-based, and about COVID-19.

### 4.3 Topic Identification in DS3

Based on our models with topic cluster numbers from three to ten (Table 8), we achieved the highest coherence scores for both fake news items and real news items when the topic cluster number was set to ***six***. Therefore, we clustered both into six topics. We generated word clouds and observed that half of the six topics overlapped between real news and fake news. These are “people and vaccine” (Figure 3a and Figure 4e), “pandemic situation in India” (Figure 3c and Figure 4d), and “state’s critical cases” (Figure 3f and Figure 4c). We derived these results by calculating the common words and similarity rates between fake news topics and real news topics. We define the similarity rate as:

**DEFINITION 2. Similarity Rate between two word clouds**

$$\text{Similarity Rate} = \frac{\text{Common Word Count in two Word Clouds}}{\text{Avg (word count in wordCloud1, word count in wordCloud2)}}$$

For example, there are 10,751 words in Figure 3a, and 9,499 words in Figure 4e. They have 6,557 words in common. Their similarity rate is 64.7%. The similarity rate of Figure 3c (6,351 words) and Figure 4d (7,054 words) is 67% (4,488 common words). The similarity rate of Figure 3f (10,418 words) and Figure 4c (11,752 words) is 66% (7,388 common words). Among 36 pairs of 6 fake news topics and 6 real news topics, these three are the only pairs that have a similarity rate above 50%.

Another finding based on our experiments is that fake news has higher coherence scores than real news for all given topic numbers (Table 8), which means that in fake news words from the same topic are more closely related and have a higher degree of semantic similarity. This implies that compared with real news, fake news might have a more consistent writing style [71].

Table 8: Coherence Scores for between 3 and 10 topic clusters of fake news and real news in DS3.

| Topics | Fake news | Real news |
|---|---|---|
| 3 | 0.58 | 0.37 |
| 4 | 0.63 | 0.29 |
| 5 | 0.57 | 0.42 |
| ***6*** | ***0.65*** | ***0.49*** |
| 7 | 0.58 | 0.45 |
| 8 | 0.56 | 0.35 |
| 9 | 0.58 | 0.44 |
| 10 | 0.55 | 0.42 |

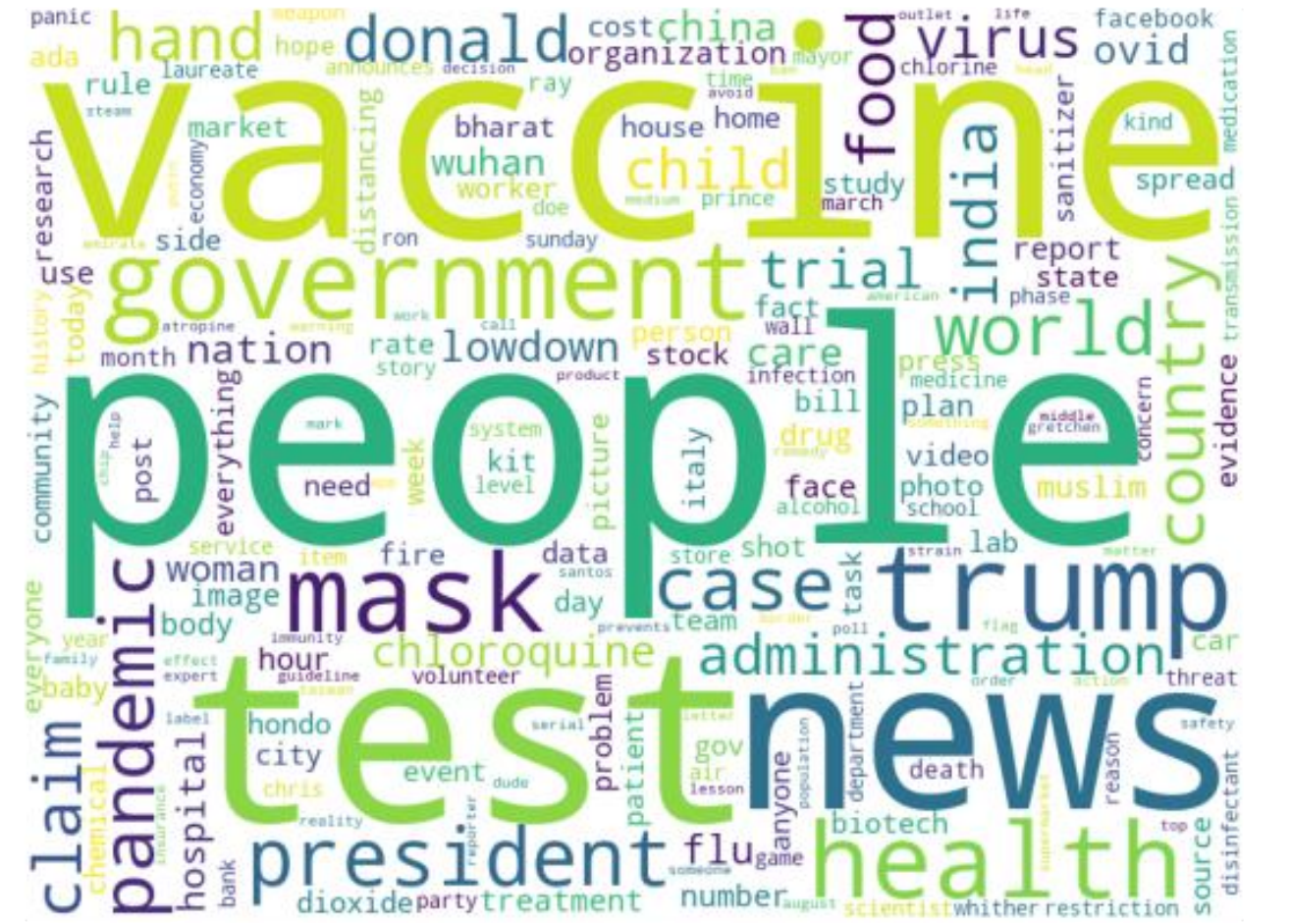

(a)

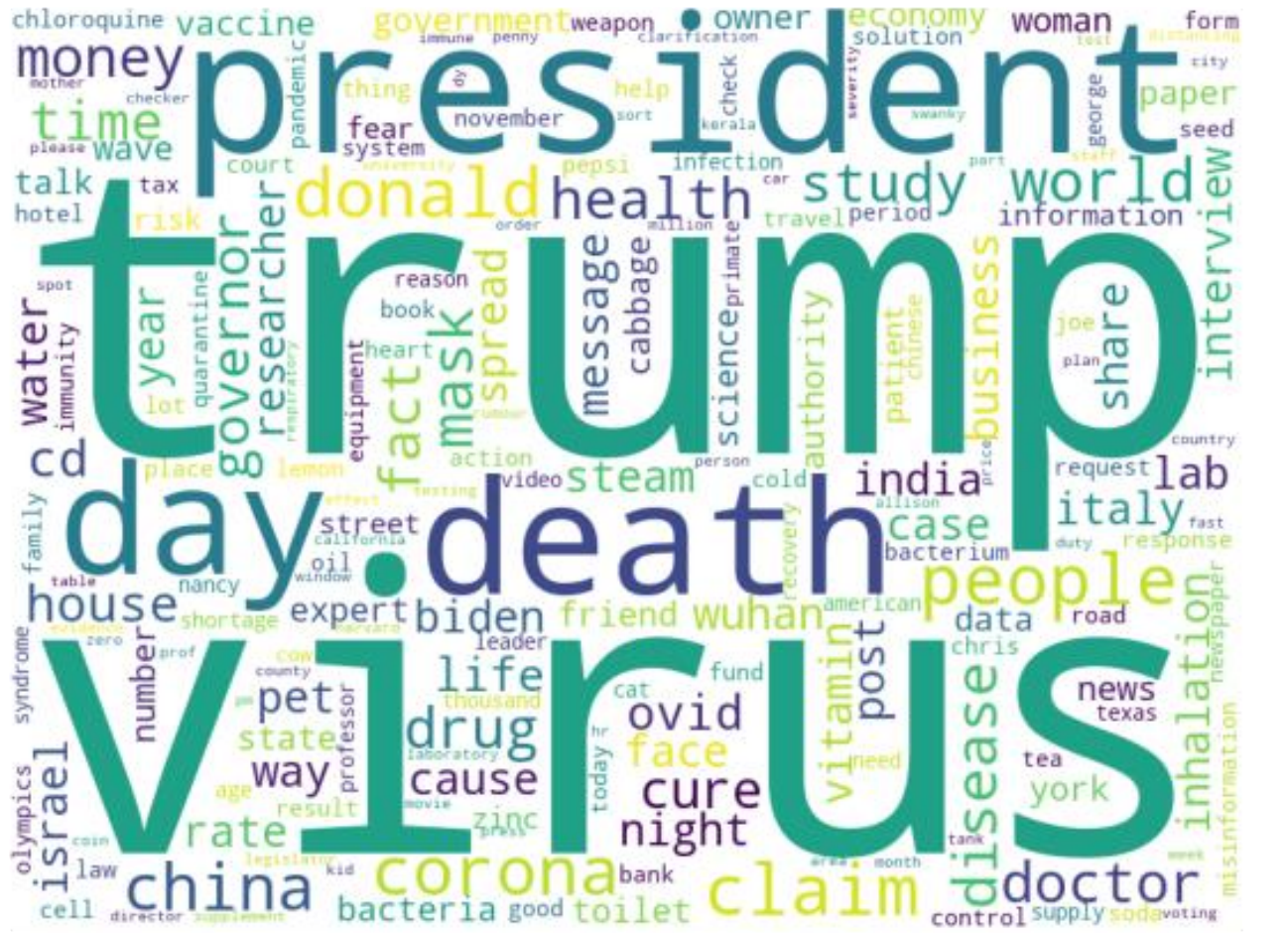

(b)

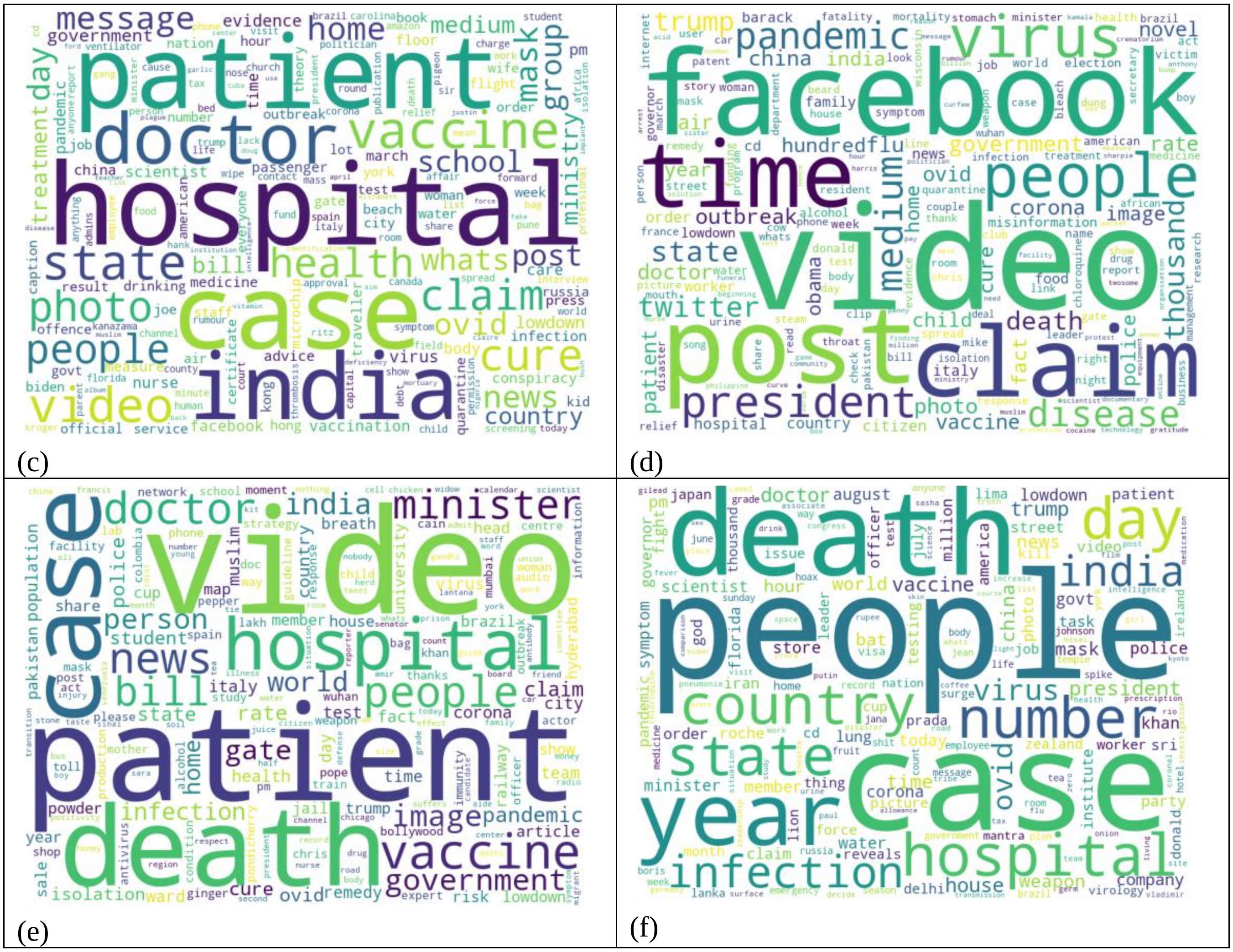

Figure 3: Word clouds generated from identified topics in **fake news** items in DS3.

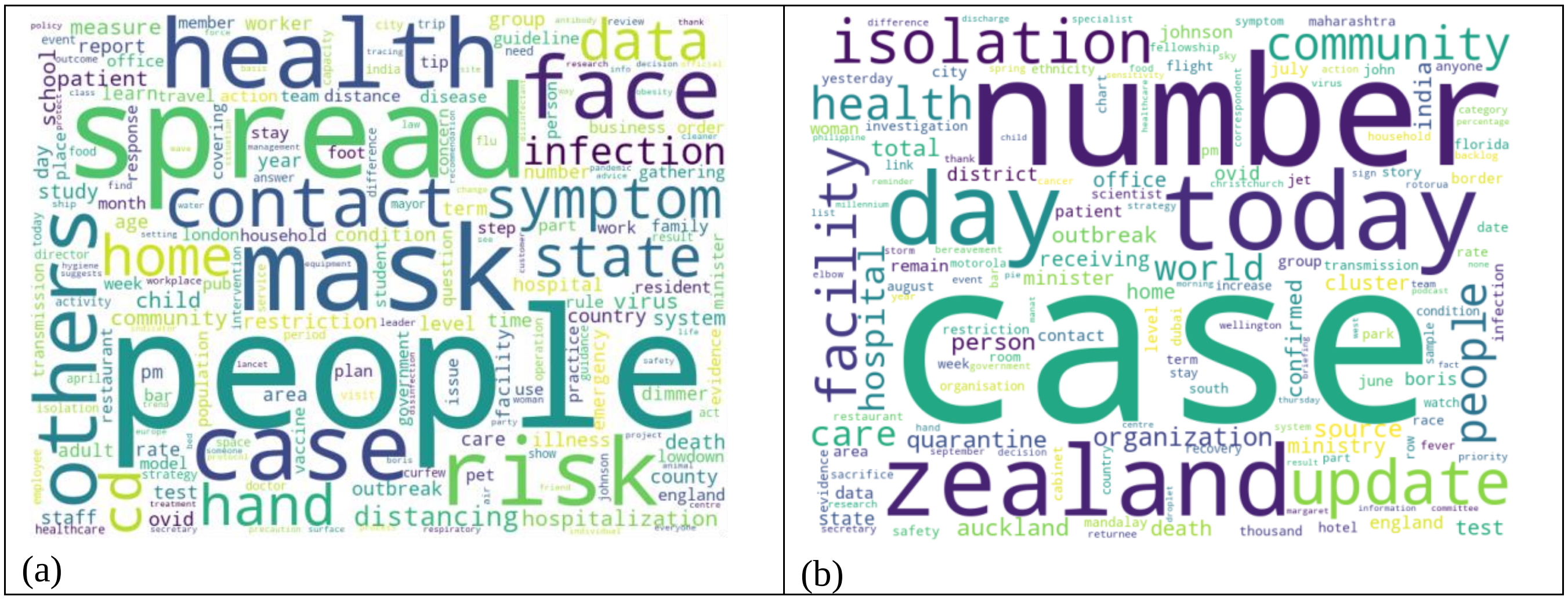

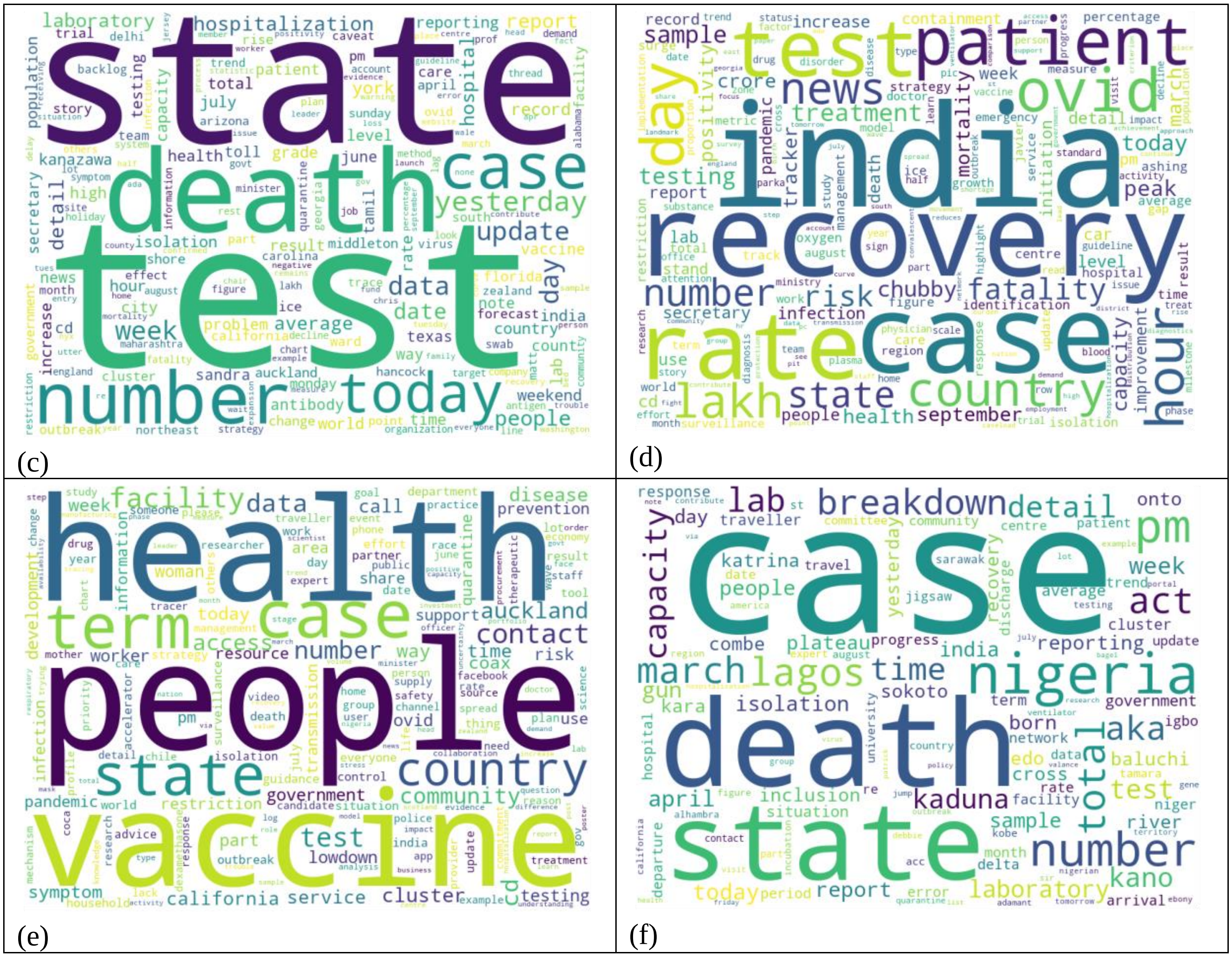

(c) (d) (e) (f)

Figure 4: Word clouds generated from identified topics in **real news** items in DS3.

### 4.4 Length of Posting as a Feature of DS3

We found that real news items are on average 40% longer than fake news items (175 char vs. 125 chars; 29.89 words vs. 20.7 words) in DS3. We hypothesize that to defend the facts, scientists/authorities might need to draft longer paragraphs of clear and precise content. For example, to combat a fake news item such as "Garlic can be used to cure COVID-19," it may be necessary to cite results from previously published studies that contradict this claim.

### 4.5 Sentiment Analysis of DS3

Table 9 presents the statistical results of sentiment analysis on DS3. The fake news items result in a higher Concern Index than real news items by 10% (72% vs. 62%).

Table 9: Statistical Results of Sentiment Analysis on DS3.

| | fake news items | real news items |
|---|---|---|
| Very Negative (VN) | 2,512 | 1,794 |
| Negative (Neg) | 247 | 553 |
| Neutral (Neu) | 240 | 677 |
| Positive (Pos) | 503 | 716 |
| Very Positive (VP) | 578 | 740 |
| Concern Index (CI) | 0.72 | 0.62 |

This 10% difference is statistically significant, based on the Z-score calculation.

**DEFINITION 3.** Z-score calculation.

$$C_F \text{ (CI of fake news) } = \frac{N_F}{T_F} = 0.72 \quad (1)$$

$$C_R \text{ (CI of real news) } = \frac{N_R}{T_R} = 0.62 \quad (2)$$

$$C_T \text{ (CI Combined)} = \frac{(N_F+N_R)}{(T_F+T_R)} = 0.67 \quad (3)$$

$$\text{std } (C_F - C_R) = sqrt\left( \left(C_T * \frac{(1-C_T)}{T_F}\right) + \left(C_T * \frac{(1-C_T)}{T_R}\right) \right) = 0.01076 \quad (4)$$

$$\text{Z-score} = \frac{abs(C_F - C_R)}{std(C_F - C_R)} = \frac{0.72-0.62}{0.01076} = \mathbf{9.3} \quad (5)$$

Where:

$N_F$ = Very Negative + Negative Fake news #,
$T_F$ = Very Negative + Negative + Very Positive + Positive Fake news #,
$N_R$ = Very Negative + Negative Real news #,
$T_R$ = Very Negative + Negative + Very Positive + Positive Real news #.

We obtained a Z-score of ***9.3.*** A lookup in Social Science Statistics (n.d.) [72] of a two-tailed p-value from the Z-score identified a ***p-value < 0.00001***. Thus, the difference of concern indices between fake news and real news is highly **significant.** This result shows that fake news expressed more negative emotions.

### 4.6 Hashtag and Mention Analysis of DS3

Figure 5 and Figure 6 show the top 30 hashtags used in fake news and real news in DS3. In fake news, there are 2,021 hashtags, 794 of which are unique, while in real news there are a total of 4,743 hashtags, 386 of which are unique. One interesting finding is the wording used for the pandemic. In fake news, **"coronavirus"** is used more often than **"covid19,"** while in real news, "**covid19**" is a more popular usage. Figure 7 and Figure 8 show the bar graphs of the top 30 hashtags that exist only in fake news vs only in real news. In fake news, hashtags tend to contain substrings such as "trump," "wuhan," "virus," "fact," "check." In real news, hashtags contain inspiring messages such as "takeresponsibility," "covidupdates," "coronaupdates," "wearamask," "slowthespread," and "reopeningsafely."

We also looked into the accounts mentioned. Figure 9 and Figure 10 present the bar graphs of the top 30 mentions in fake news and in real news. In fake news, there are 669 mentions, 486 of which are unique, while in real news, there are 2,090 mentions, 568 of which are unique. In fake news, the top mentions are often politically related: "realDonalTrump," "narandramodi" (Indian Prime Minister), "factchecknet," "PMOIndia" (Prime Minister Office India), while in real news, top mentions are related to public health experts or institutes: "MoHFW_INDIA" (Ministry of Health and Family Welfare of India), "ICMRDELHI" (India Council of Medical Research), "DrTedros" (Director General of WHO), and "drharshvardhan" (An Indian Otorhinolaryngologist).

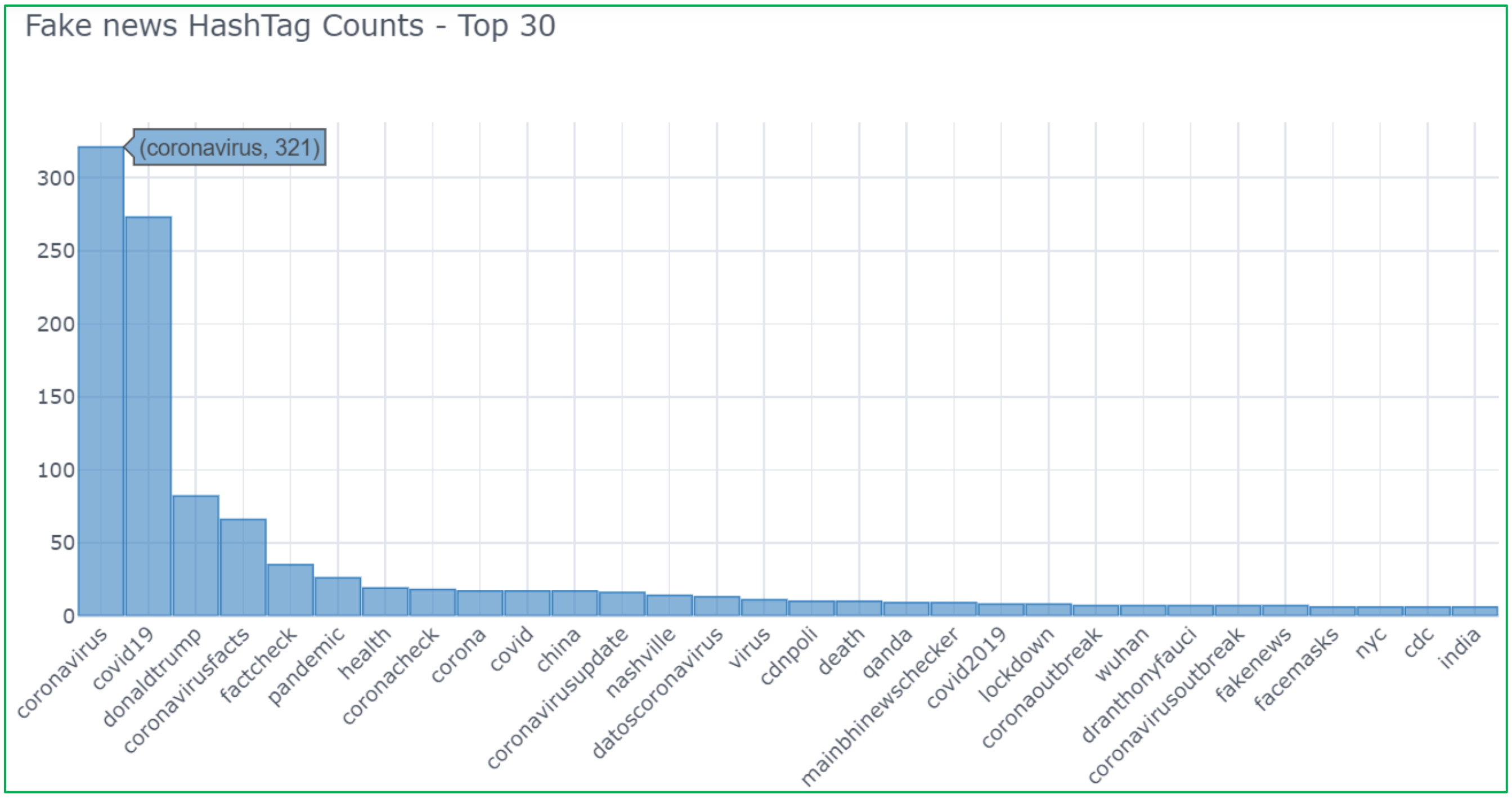

Figure 5: Bar graph showing top 30 hashtags used in fake news.

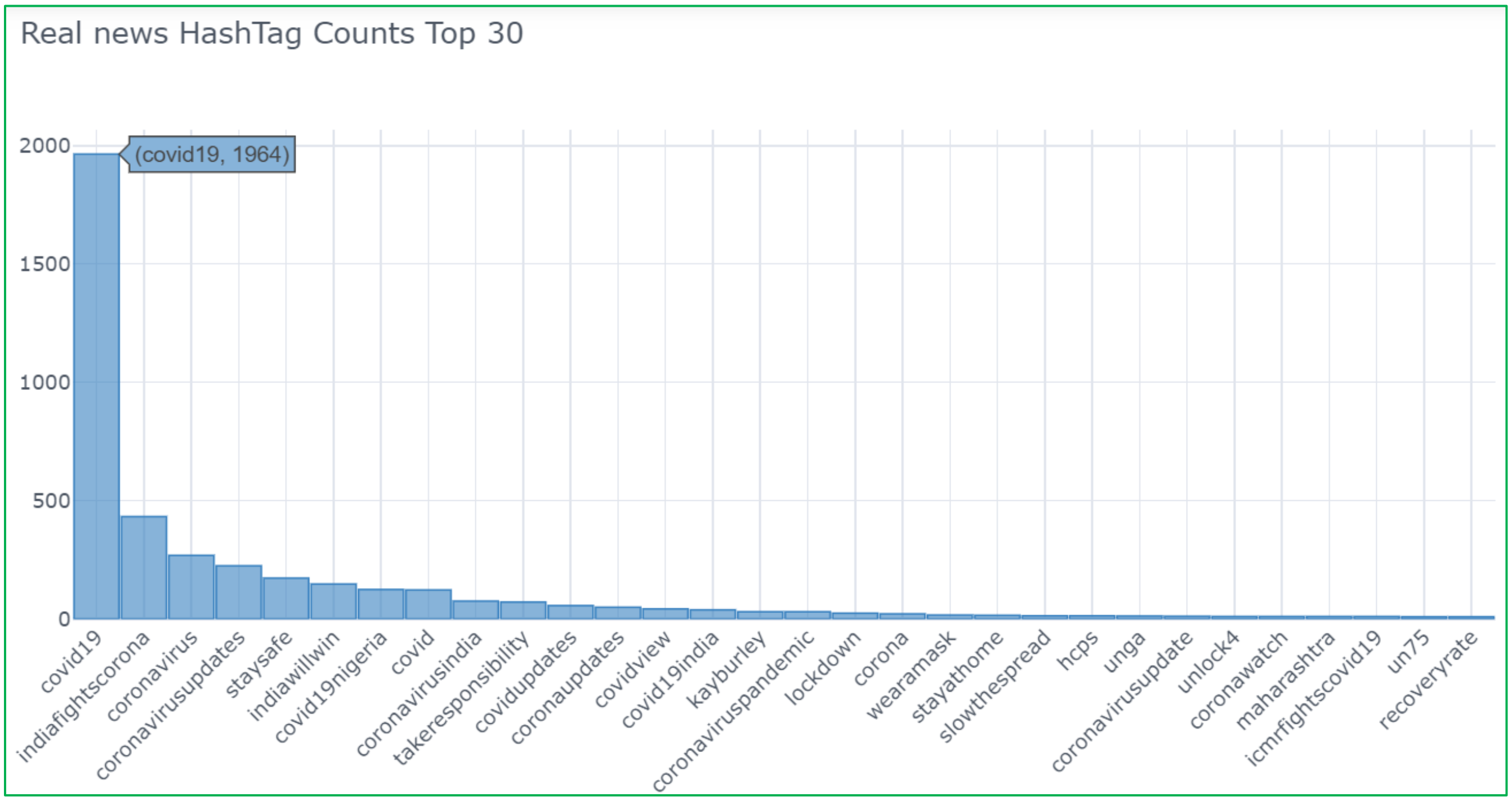

Figure 6: Bar graph showing top 30 hashtags used in real news.

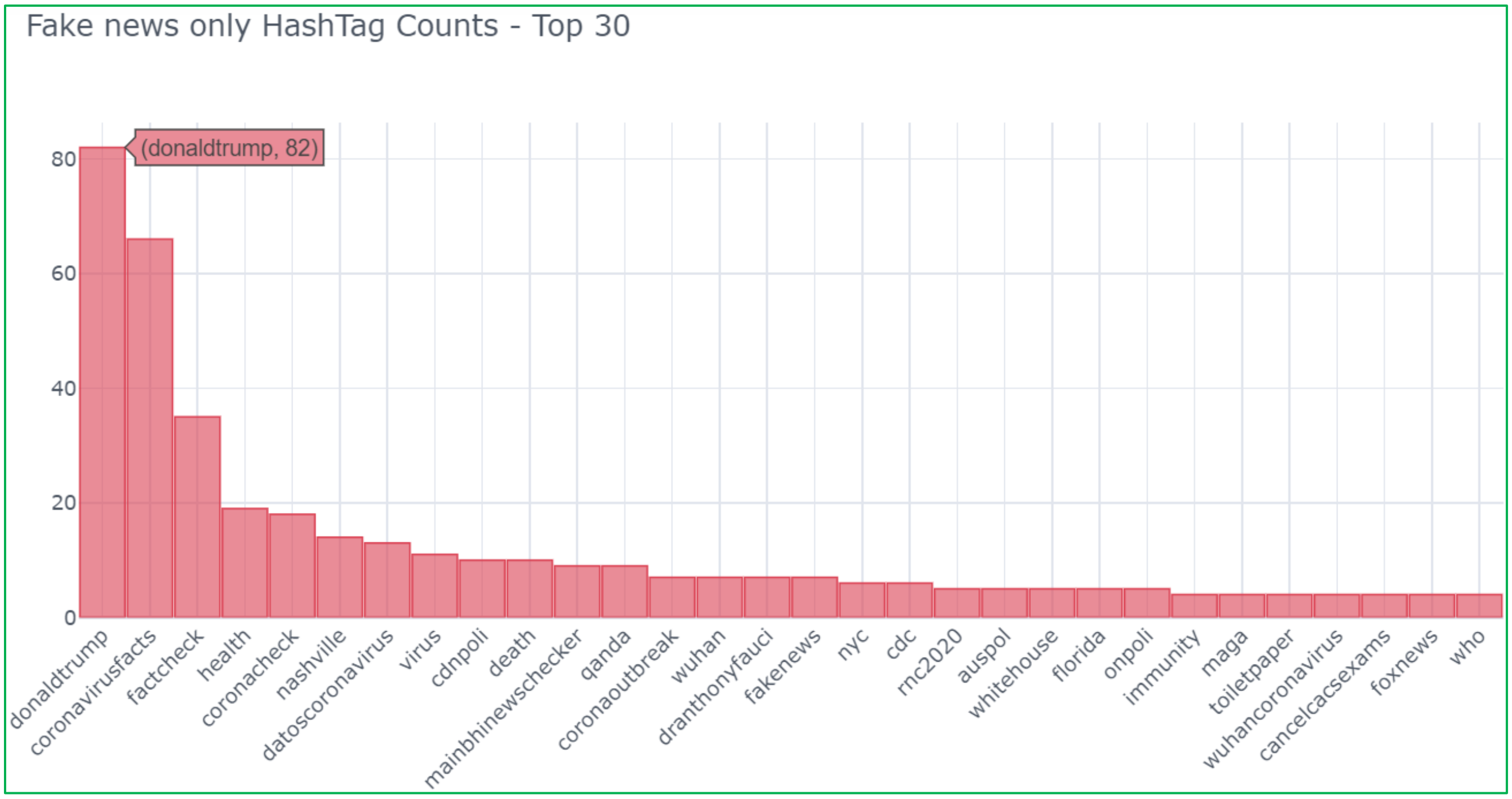

Figure 7: Bar graph of top 30 hashtags occurring only in fake news.

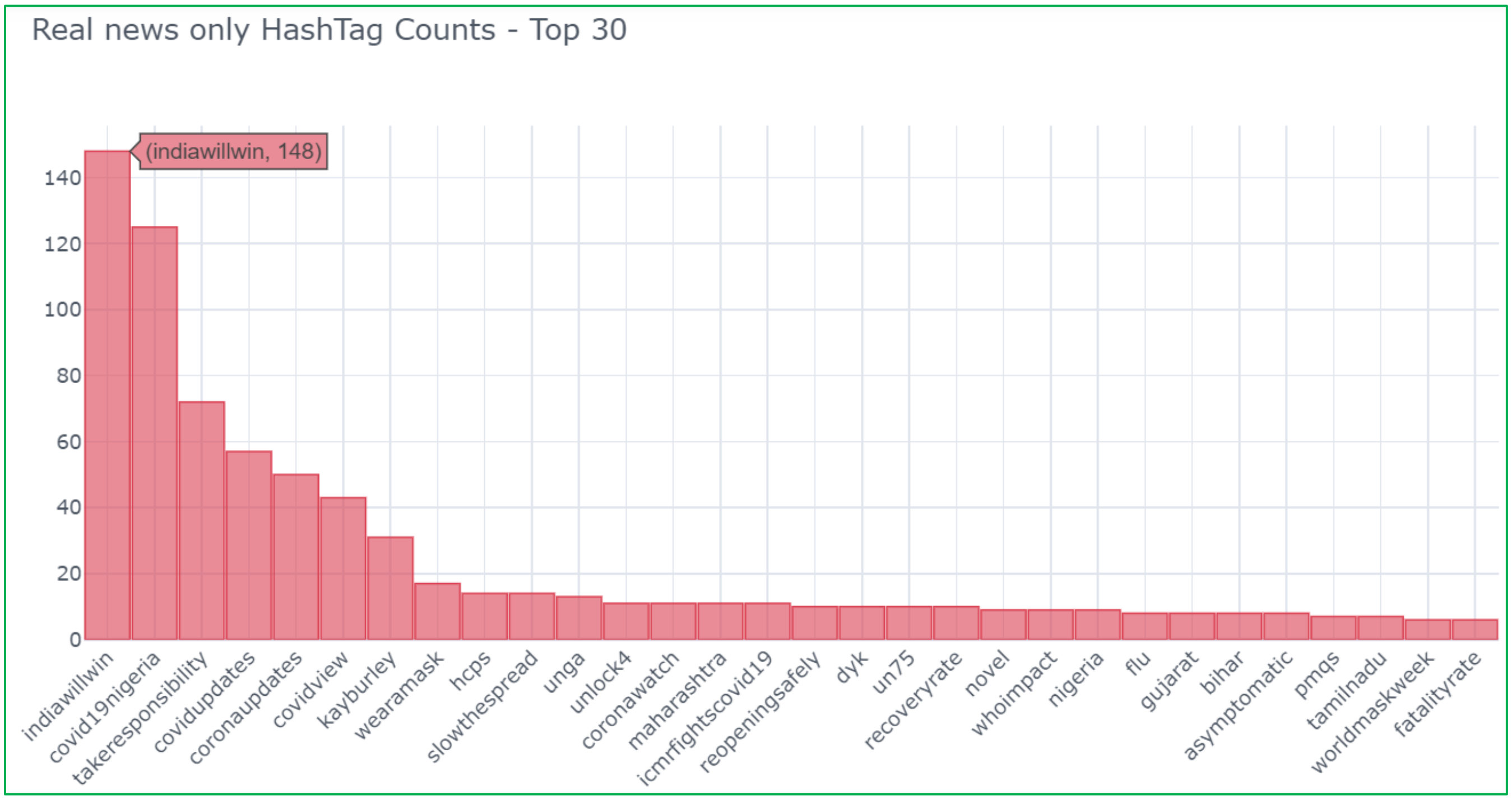

Figure 8: Bar graph showing top 30 hashtags occurring only in real news.

We also did a normalization experiment, based on the count difference between fake news (4,080) and real news (4,480). Since the dataset is class-wise balanced, the comparison results between fake news and real news discussed above did not change.

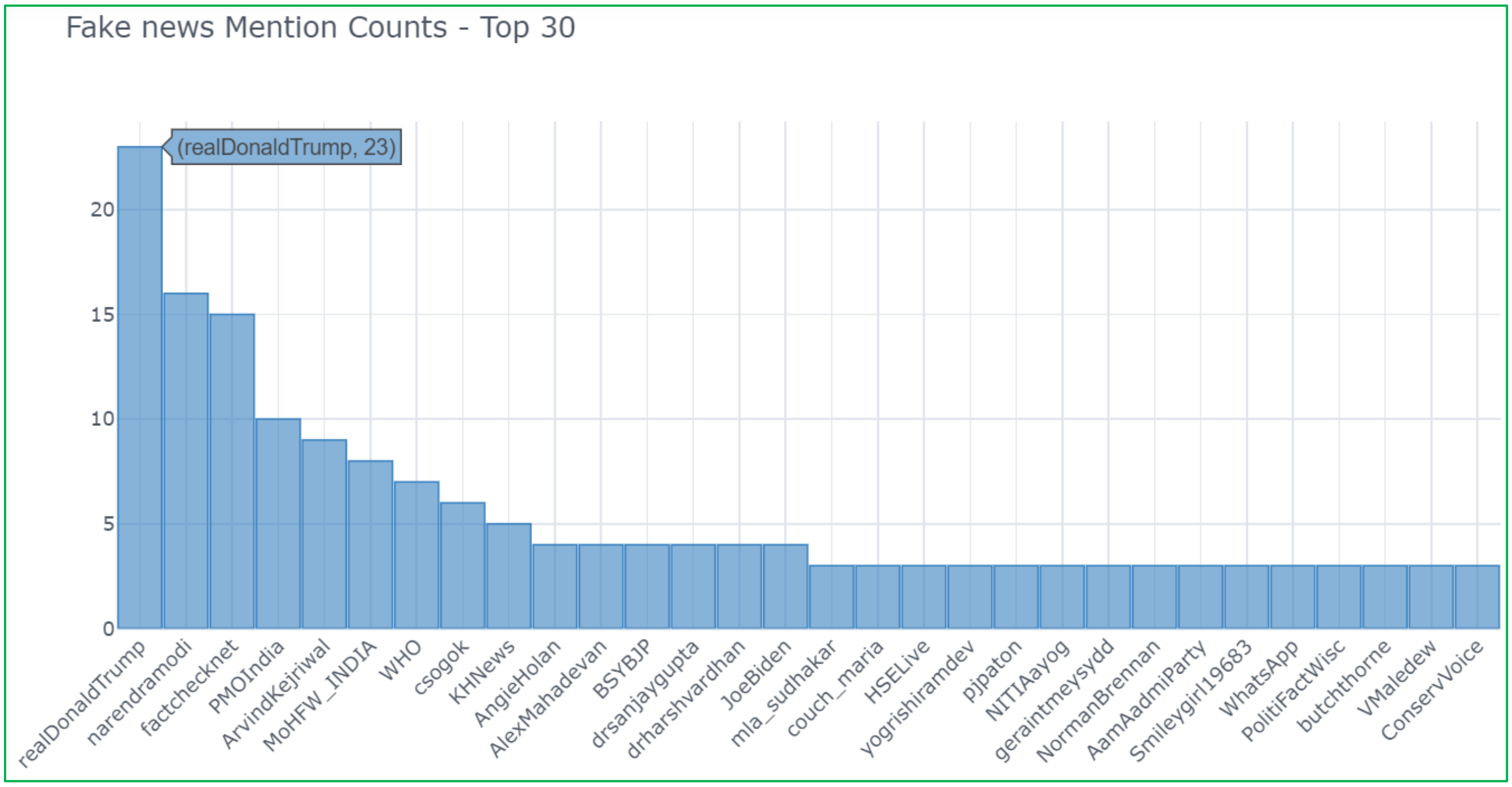

Figure 9: Bar graph showing top 30 mentions in fake news.

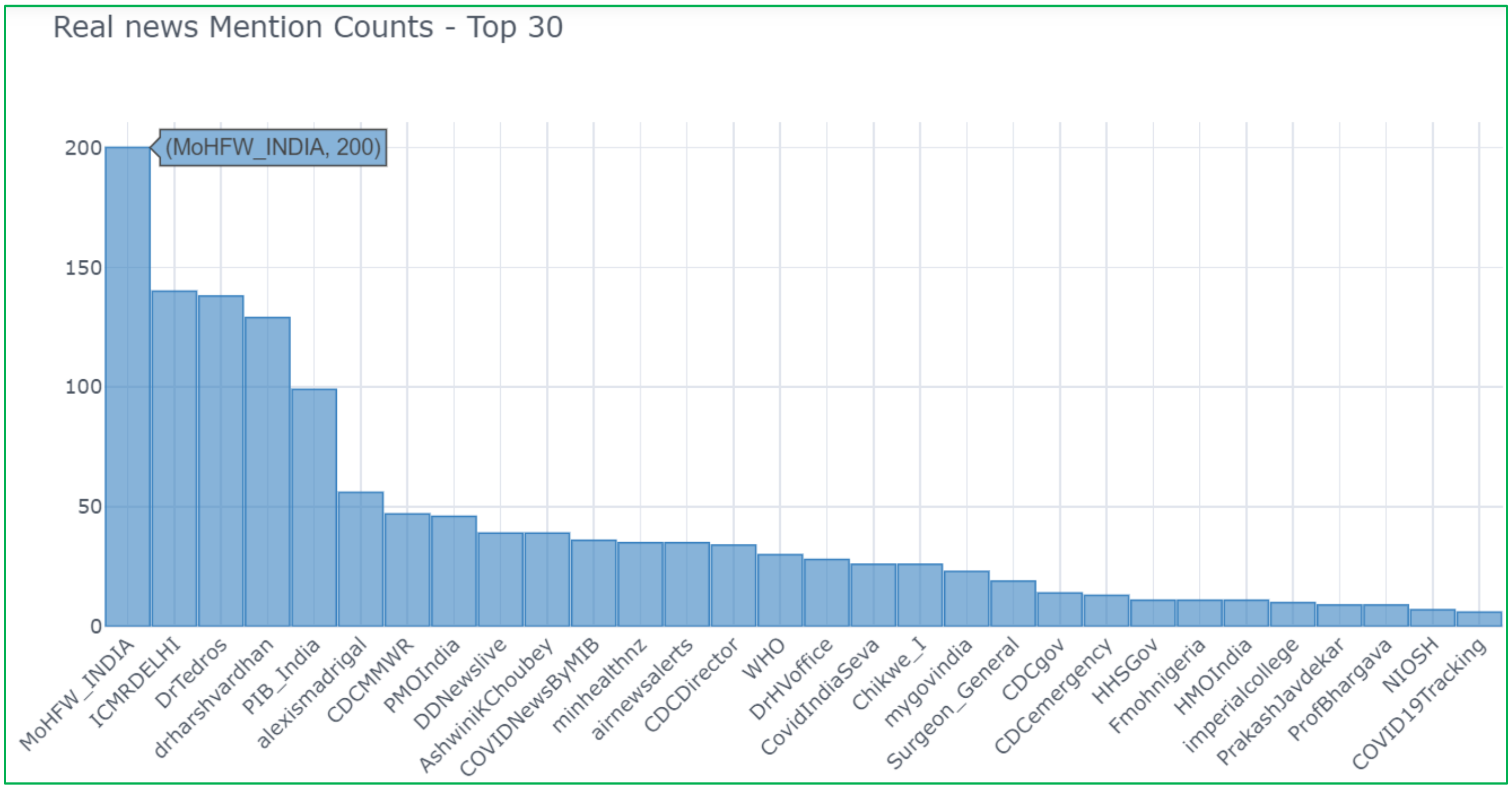

Figure 10: Bar graph showing top 30 mentions in real news.

### 4.7 BERT models for DS3 with feature elimination

We also investigated how influential the behavioral features are for fake news detection. We trained BERT models on DS3 with feature elimination. Namely, we performed sensitivity analysis [73] and tested model performance. The selected features included hashtags and mentions, and the feature elimination consisted of (1) remove hashtags, but keep mentions, (2) keep hashtags, but remove mentions, and (3) remove both hashtags and mentions. To be precise, removing a hashtag means that the word that is preceded by the character “#” is completely removed, as opposed to leaving the word and removing only the “#” from it. Out of 8,560 data rows in DS3, there are 286 rows containing mentions, and 648 rows containing hashtags. A finding here is that in DS3 there is no data row containing both a mention and a hashtag.

We compared the accuracy among the models trained on original text, and text with feature(s) eliminated. As shown in Table 10, the achieved accuracy were slightly different for different combinations. The model trained on the original text without feature elimination performed best (as already presented in Table 6), the model trained on the text with both features eliminated had the lowest accuracy, and the models trained on text with either feature eliminated had accuracy in between. Though the uses of hashtags and mentions differ between fake news and real news, the influence of hashtags and mentions towards fake news detection is not significant. This might be due to the low percentage of data rows containing hashtags or mentions.

Table 10: BERT Model Accuracy on DS3 with feature elimination.

| Original Text | Eliminate Hashtag | Eliminate Mention | Eliminate both |
|---|---|---|---|
| **95.61%** | 95.03% | 95.41% | 94.81% |

### 4.8 BERT models for DS3 with features added into sentence embedding.

We will use the term “identified features” to include the behavioral features and the word and character counts. To investigate how the identified features (i.e., sentiment, number of words, number of characters, number of hashtags, and number of mentions) of DS3 would influence the fake news detection, we added them (in numerical format) into the sentence embeddings used for BERT model training. During sentence embedding of BERT, each sentence processed by BERT is tokenized, and each token is converted into an integer id. BERT itself has a dictionary with 30,254 rows, and each row contains exactly one token. The id of a token is its row number minus one.

To avoid duplicate ids in sentence embeddings, we used the sum of 31,000 and the sentiment value of the sentence as the token id of the sentiment. For the sentiments themselves, we converted “Very negative” and “Negative” to 0, “Neutral” to 2, and “Very Positive” and “Positive” to 4. Correspondingly, 32,000 is added to the number of words in the sentence as the token id of the word count. In a similar manner, 33,000 is used to offset the character count from the token ids used by BERT. Next, 34,000 is added to the hashtag count and 35,000 is added for the mentions count.

For example, the fake news item in DS3 “*Washington Examiner Editor Loses Head Up His Ass #washington #josephbiden #covid19*” is encoded as follows. This sentence expressed a negative sentiment. The token id of its sentiment is 31,000 (0+31,000). The token id of its word count is 32,011 (11 words + 32,000). The token id of its character count is 33,082 (82 chars + 33,000). The token id of its hashtag count is 34,003 (3 hashtags + 34,000). The token id of its mention count is 35,000 (0 mentions + 35,000). This is shown in Figure 11. BERT did a sentence embedding and converted input text into numerical data, stored in a list. We appended the converted numerical data of the five features to the list. The token ids 101 and 102 are ids of special tokens “[CLS]” and “[SEP]” for BERT to represent the input properly.

Washington Examiner Editor Loses Head Up His Ass #washington #josephbiden #covid19

token ids of BERT sentence embedding**:** **[**101**,** 2899, 19684, 3559, 12386, 2132, 2039, 2010, 4632, 1001, 2899, 1001, 3312, 17062, 2368, 1001, 2522, 17258, 16147, 102**]**

tokens id of BERT sentence embedding **after we added the identified features**: **[**101, 2899, 19684, 3559, 12386, 2132, 2039, 2010, 4632, 1001, 2899, 1001, 3312, 17062, 2368, 1001, 2522, 17258, 16147, **31000, 32011, 33082, 34003, 35000,** 102**]**

Figure 11: Sentence Embedding of BERT before and after we added the identified features.

We then built BERT detection models with the added features, and performed feature elimination as in the previous subsection. The results are shown in Table 11. The column header “Text with Hashtag and Mention” means that we used the sentence embedding with five features appended. “Text with Hashtag Eliminated” means the hashtags were deleted from the post set. The same applies to “Text with Mention eliminated.” In those cases, the appeded numbers default to 34,000 and 35,000. The accuracy of models with the identified features added into the sentence embeddings do not reflect improvements over the models without the additional features.

Table 11: BERT Model Accuracy on DS3 with added features, and feature elimination.

| Text with Hashtag and Mention | Text with Hashtag Eliminated | Text with Mention eliminated | Text with Hashtag and Mention Eliminated |
|---|---|---|---|
| **94.72%** | 93.91% | 94.36% | 92.22% |

### 4.9 Combination of Models

This subsection is an extension to the previous subsection. We combined the Deep Learning model BERT with SVM, and used both plain text and derived features in an ensemble model to improve model robustness. For SVM we used numerical features instead of the text for fake news detection based on DS3. Namely, SVM used the derived features of each post including sentiment, number of words, number of characters, number of hashtags, and number of mentions. The best accuracy of the SVM model by itself is 80.02%, when the hyperparameters of the SVM model are set to C = 1000.0, kernel = “rbf,” and gamma = “auto.”

We obtained an accuracy of 97% by combining BERT and SVM, which is an improvement over BERT alone. Note that this is neither the result of a majority vote nor of a feature combination. Rather, first, out of 8,560 data items in DS3, BERT and SVM predicted the same label (0 or 1) in 6,705 cases. Of these, 6,514 were correct (according to the training data). Therefore, the accuracy of the ensemble model is 6,514/6,705 = 97%.

### 4.10 Statistics Regarding DS3

We present (in Table 12) a summary of the statistics of DS3. The table includes post counts, average word counts, average character counts, training/test split, sentiment distribution, total hashtag counts, unique hashtag counts, total mention counts and unique mention counts.

Table 12: Statistics Regarding DS3.

| | | fake news items | real news items |
|---|---|---|---|
| Counts and Length of Posts | Post Counts | 4,080 | 4,480 |
| | AVG Word Counts per post | 20.7 | 29.89 |
| | AVG Char Counts per post | 125 | 175 |
| Training/Test Split | | 80% for training and 20% for testing.<br>5-fold cross validation. | |
| Senti-ment Distribu-tion | Very Negative | 2,512 | 1,794 |
| | Negative | 247 | 553 |
| | Neutral | 240 | 677 |
| | Positive | 503 | 716 |
| | Very Positive | 578 | 740 |
| Hashtags | | Totally 2,021,<br>794 out of which are unique. | Totally 4,743,<br>386 out of which are unique. |
| Mentions | | Totally 669,<br>486 out of which are unique. | Totally 2,090,<br>568 out of which are unique. |

## 5. Discussion

Fake news circulating on social media, no matter whether created by mistake or intention, has created trust issues among citizens and discord in society. We built Deep Learning models for fake news detection based on different domains of datasets. While our BERT models achieved state-of-the-art results compared with previous studies, we found that the models did not do well when evaluated on the other two datasets. To detect COVID-related fake news with models trained with non-COVID datasets, appropriate adjustments for transfer learning will be required. The DS2 dataset was constructed for this research, and thus no comparison with published research is possible. Among all our experiments, the BERT2 model showed the best performance.

We built topic identification models to identify topics in both fake news and real news in DS3. We found that half of the identified topics were overlapping across the news items. Such overlap of news topics across real news and fake news can add difficulties for citizens to distinguish between fake news and real news. We performed behavioral and sentiment analysis on DS3 and identified a number of feature differences between fake news items and real news items. The features include length of post, Concern Index, use of hashtags, and use of mentions. We hypothesized that such differences could help distinguish fake news from real news. For this purpose, we built BERT detection models for DS3 with the features (in numerical format) concatenated into the sentence embeddings. Unfortunately, we obtained slightly lower accuracy than for models built without the features added into the sentence embeddings. As BERT was pre-trained using plain text only, additional features in numerical data might not help improve model accuracy. Besides, the numbers applied were out-of-vocabulary (OOV) for BERT's sentence embeddings. Therefore, the numbers in the embeddings were not meaningful, and could even be interpreted as noise by BERT.

One result that challenged our intuitions was that fake news achieved higher coherence scores than real news. A possible explanation might be that real news will often be supported by a "proof," while fake news spreaders often try convincing the world by simple language that is often repeated.

## 6. Conclusions

In this paper, we built Deep Learning models for fake news detection based on different domains of datasets. Our BERT models have better performance than previous studies. By combining the BERT Deep Learning model with SVM, we improved the ensemble model's accuracy based on DS3.

To further realize the features of fake news posts generated and shared on social media amid the COVID-19 pandemic, we applied a number of NLP techniques to DS3. We discovered a number of interesting findings that are also summarized in Table 13:

- The Concern Index of fake news is great than that of real news by 10%. Based on the derived p-value, this is a significant difference. This result answered Question 1 raised in the Introduction: fake news expressed more negative emotions, and the difference is **significant**, which shows that fake news can cause negative emotional impacts on citizens and in society.
- Real news posts are on average 40% longer than fake news. This implies that to recognize fake news, length can provide a hint. Regarding the use and count of hashtags, fake news contains more unique hashtags while real news has more total hashtags. Fake news posters tend to use "coronavirus" to describe the pandemic, while real news users tend to write "covid19." In fake news, hashtags tend to contain substrings such as "trump," "wuhan," "virus," "fact," or "check," while in real news, hashtags contain inspiring messages such as "indiawillwin," "takeresponsibility," "covidupdates," "coronaupdates," "wearamask," "slowthespread," "icmfightscovid19," and "reopeningsafely." When it comes to the count of mentions, real news contains more unique mentions and more total mentions than fake news. In fake news, top mentions are the handles of politicians and fact check sites, while in real news, top mentions are the handles of public health experts and institutes. To further realize whether the identified features of hashtags and mentions help achieve better performance in fake news detection, we built detection models based on feature elimination. The detection model with both features achieved the best accuracy, while the model with both features eliminated showed the worst performance. The findings presented in this paragraph provide an answer to Question 2 in the Introduction.
- By using the best coherence score, based on topic numbers between three and ten, we identified six topics each for fake news and real news. We found that three of the identified topics are common for fake news and real news, namely "people and vaccine," "pandemic situation in India," and "state's critical cases." We also found that fake news has a higher coherence score than real news for our data set, which shows that fake news seems to have a more consistent writing style. These results address Question 3 from the Introduction.
- Due to the different nature of datasets, transfer learning with models trained with datasets that

were collected before COVID-19 to a COVID-19-related dataset did not work well. Based on our experiment, adjustments for transfer learning will be required to detect fake news across different domains of datasets. This result answers Question 4 in the Introduction.

- After we added the five extracted features into the sentence embeddings, the detection model performance was not improved. This finding answers Question 5 in the Introduction.

Table 13: Experimental Results and Comparison of fake news and real news in DS3.

| | Fake news | Real news |
|---|---|---|
| Hashtag | More unique hashtags. | More total hashtags. Inspiring and admonishing messages are expressed. |
| Mention | Top mentions are Twitter handles of politicians and fact checking sites. | More unique and total mentions. Top mentions are Twitter handles of public health experts and institutes. |
| Length | Real news posts are on average 40% longer than fake news. This implies that to recognize fake news, length can provide a hint. | |
| Concern Index | The Concern Index of fake news is greater than that of real news by 10%. This is a statistically significant difference. | |
| Topic Identification | Six topics were identified in both kinds of news items, half of which were overlapping, including “people and vaccine,” “pandemic situation in India,” and “state’s critical cases.” A higher coherence score shows a relatively consistent writing style of fake news. | |

## 7. Future Work

We are working on achieving better transferability for fake news data from different domains, and building detection models that can identify cross-domain fake news. Based on our detection models, we will build a fact-checking web platform, which will help citizens get an immediate response regarding the authenticity of a news item that is copied and pasted into a window of a web page.

The eventual goal of these methods is to serve social media platforms in an invisible manner. By including these Machine Learning models, social media platforms can decide to reject any messages that are recognized as fake news. Alternatively, they can let suspicious messages pass through to the user community, but provide a pop-up warning to readers that the message is potentially fake news. The second solution would be more acceptable to activists who are concerned about free speech in social media [74].

We are also conducting semantic analysis of fake news post content by knowledge graphs to further uncover semantic relations among entities in related posts. Furthermore, we will summarize fake news posts and activities based on the large number of relations and entities in fake news data.